\documentclass[10pt,twocolumn,letterpaper]{article}

\usepackage[pagenumbers]{cvpr} 
\newcommand\blfootnote[1]{%
  \begingroup
  \renewcommand\thefootnote{}\footnote{#1}%
  \addtocounter{footnote}{-1}%
  \endgroup
}

\usepackage[dvipsnames]{xcolor}

\definecolor{cvprblue}{rgb}{0.21,0.49,0.74}
\definecolor{cvprgreen}{rgb}{0.10, 0.52, 0.27}
\definecolor{cvprgrey}{rgb}{0.5, 0.52, 0.5}
\definecolor{darkblue}{RGB}{0,0,180} 
\definecolor{darkred}{RGB}{180,0,0} 
\definecolor{darkpurple}{RGB}{120,0,180} 
\definecolor{darkgreen}{RGB}{0,120,0} 
\definecolor{darkbrown}{RGB}{100,40,0} 
\definecolor{grey}{RGB}{128,128,128} 
\usepackage[pagebackref,breaklinks,colorlinks,citecolor=cvprblue]{hyperref}

\def\paperID{12749} 
\def\confName{CVPR}
\def\confYear{2024}

\title{Transcending Forgery Specificity with Latent Space Augmentation \\ for Generalizable Deepfake Detection}

\author{
    Zhiyuan Yan$^{1}$ \quad Yuhao Luo$^{1}$ \quad Siwei Lyu$^{2}$ \quad Qingshan Liu$^{3}$ \quad Baoyuan Wu$^{1, \dagger}$\\
    $^{1}$The Chinese University of Hong Kong, Shenzhen (CUHK-Shenzhen), China\\
    $^{2}$University at Buffalo, State University of New York, USA\\
    $^{3}$Nanjing University of Information Science and Technology, China\\
    {\tt\small yanzhiyuan1114@gmail.com, luo7502@gmail.com}\\
    {\tt\small siweilyu@buffalo.edu, qsliu@nuist.edu.cn, wubaoyuan@cuhk.edu.cn}
}

\begin{document}
\maketitle
\begin{abstract}
Deepfake detection faces a critical generalization hurdle, with performance deteriorating when there is a mismatch between the distributions of training and testing data.
A broadly received explanation is the tendency of these detectors to be overfitted to forgery-specific artifacts, rather than learning features that are widely applicable across various forgeries.
To address this issue, we propose a simple yet effective detector called LSDA (\underline{L}atent \underline{S}pace \underline{D}ata \underline{A}ugmentation), which is based on a heuristic idea: representations with a wider variety of forgeries should be able to learn a more generalizable decision boundary, thereby mitigating the overfitting of method-specific features (see Fig.~\ref{fig:toy}).
Following this idea, we propose to enlarge the forgery space by constructing and simulating variations within and across forgery features in the latent space.
This approach encompasses the acquisition of enriched, domain-specific features and the facilitation of smoother transitions between different forgery types, effectively bridging domain gaps.
Our approach culminates in refining a binary classifier that leverages the distilled knowledge from the enhanced features, striving for a generalizable deepfake detector.
Comprehensive experiments show that our proposed method is surprisingly effective and transcends state-of-the-art detectors across several widely used benchmarks.
\end{abstract}

\blfootnote{$^\dagger$Corresponding Author}

\section{Introduction}

\label{sec:intro}

Deepfake technology has rapidly gained prominence due to its capacity to produce strikingly realistic visual content. Unfortunately, this technology can also be used for malicious purposes, \textit{e.g.,} infringing upon personal privacy, spreading misinformation, and eroding trust in digital media.  Given these implications, there is an exigent need to devise a reliable deepfake detection system.

\begin{figure}[!t] 
\centering 
\includegraphics[width=0.478\textwidth]{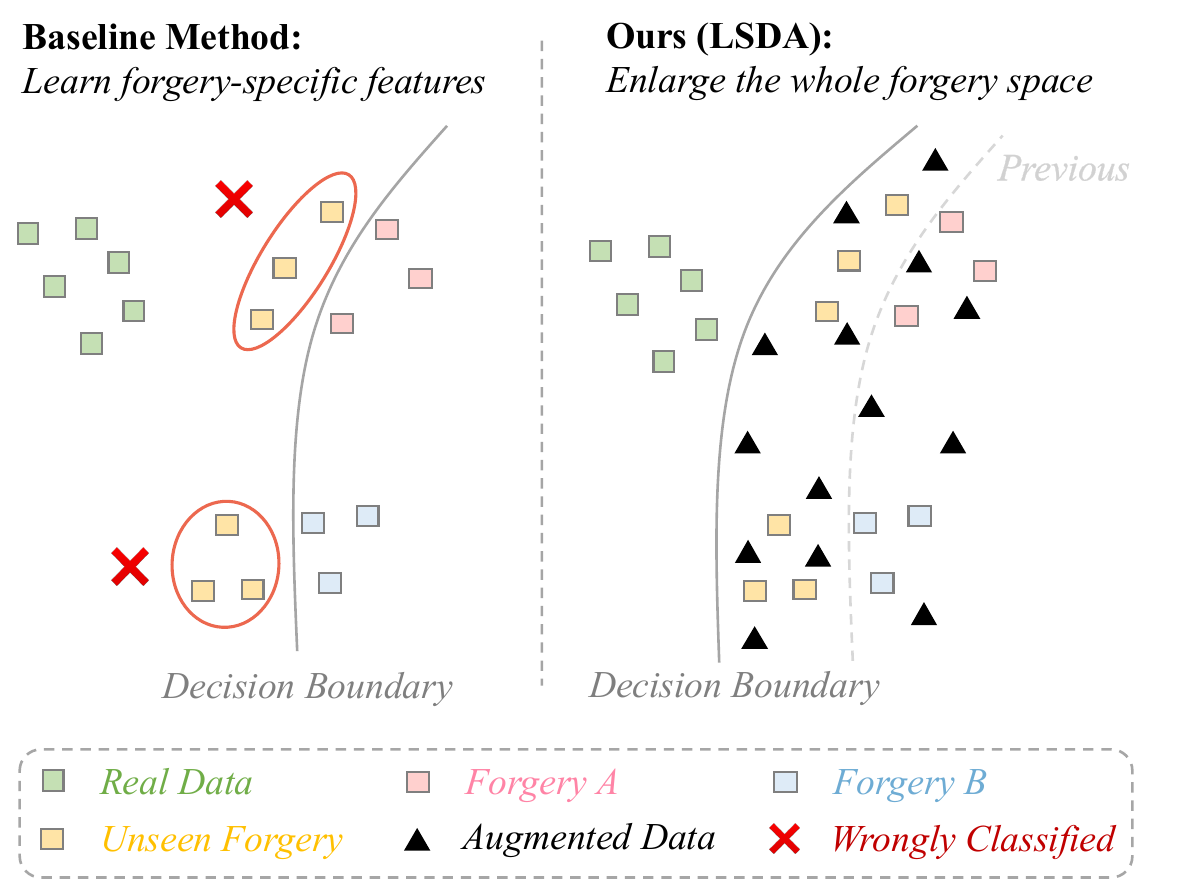} 
\caption{
Toy examples for intuitively illustrating our proposed latent space augmentation strategy. The baseline can be overfitted to forgery-specific features and thus cannot generalize well for unseen forgeries. In contrast, our proposed method avoids overfitting to specific forgery features by enlarging the forgery space through latent space augmentation. This approach aims to equip our method with the capability to effectively adjust and adapt to new and previously unseen forgeries. 
} 
\label{fig:toy} 
\end{figure}

The majority of previous deepfake detectors~\cite{zhou2017two,li2018exposing,rossler2019faceforensics++,sabir2019recurrent,yang219exposing,qian2020thinking,zhao2021multi} exhibit effectiveness on the within-dataset scenario, but they often struggle on the cross-dataset scenario where there is a disparity between the distribution of the training and testing data.
In real-world situations characterized by unpredictability and complexity, one of the most critical measures for a reliable and efficient detector is the generalization ability. However, given that each forgery method typically possesses its specific characteristics, the overfitting to a particular type of forgery may impede the model's ability to generalize effectively to other types (also indicated in previous works~\cite{luo2021generalizing,shiohara2022detecting,yan2023ucf}).


In this paper, we address the generalization problem of deepfake detection from a heuristic idea: \textit{enlarging the forgery space through interpolating samples encourages models to learn a more robust decision boundary and helps alleviate the forgery-specific overfitting}. We visually demonstrate our idea in Fig.~\ref{fig:toy}, providing an intuitive understanding.
Specifically, to learn a comprehensive representation of the forgery, we design several tailored augmentation methods both within and across domains in the latent space. For the within-domain augmentation, our approach involves diversifying each forgery type by interpolating challenging examples\footnote{Challenging examples are that farthest from the center. Within each mini-batch, they are determined by measuring the Euclidean distance between the mean of the samples and other samples.}. The rationale behind this approach is that challenging examples expand the space within each forgery domain. 
For the cross-domain augmentation, we utilize the effective Mixup augmentation technique~\cite{zhang2017mixup} to facilitate smooth transitions between different types of forgeries by interpolating latent vectors with distinct forgery features. 

Moreover, inspired by previous work~\cite{haliassos2022leveraging}, we leverage the pre-trained face recognition model ArcFace~\cite{deng2019arcface} to help the detection model learn a more robust and comprehensive representation for the real. It is reasonable to believe that the pre-trained face recognition model has already captured comprehensive features for real-world faces. Therefore, we can employ these learned features to finetune our classifier to learn features of the real.
Our approach culminates in refining a binary classification model that leverages the distilled knowledge from the comprehensive forgery and the real features. In this manner, we aim to strive for a more generalizable deepfake detector. 


Our proposed latent space method offers the following potential advantages compared to other RGB-based augmentations~\cite{li2018exposing,li2020face,zhao2021learning,chen2022self}.
\textbf{Robustness}: these RGB-based methods typically synthesize new face forgeries (pseudo fake) through pixel-level blending to reproduce simulated artifacts, \textit{e.g.,} blending artifacts~\cite{li2020face,zhao2021learning}. However, these artifacts could be susceptible to alterations caused by post-processing steps, such as compression and blurring (as verified in Fig.~\ref{fig:dongge}). 
In contrast, since our proposed augmentation only operates in the latent space, it does not directly produce and rely on pixel-level artifacts for detection.
\textbf{Extensibility}: these RGB-based methods typically rely on 
some specific artifacts (\eg, blending artifacts), which may have limitations in detecting entire face synthesis~\cite{karras2019style} (as verified in Tab.~\ref{tab:detect_generative}). This limitation stems from the fact that these methods typically define a ``fake image" as one in which the face-swapping operation (blending artifact) is present.
In contrast, our method aims to perform augmentations in the latent space that do not explicitly depend on these specific pixel-level artifacts for detection.


Our experimental studies confirm the effectiveness of our proposed method. We surprisingly observe a substantial improvement over the baseline methods within the deepfake benchmark~\cite{yan2023deepfakebench}. Moreover, our method demonstrates enhanced generalization and robustness in the context of cross-dataset generalization, favorably outperforming recent state-of-the-art detectors.

\section{Related Work}
\label{sec:related}

\paragraph{Deepfake Generation Methods}
Deepfake generation typically involves face-replacement~\cite{fs,df,li2019celeb}, face-reenactment~\cite{thies2016face2face,thies2019deferred}, and entire image synthesis~\cite{karras2017progressive,karras2019style}. Face-replacement generally involves the ID swapping utilizing the auto-encoder-based~\cite{df,li2019celeb} or graphics-based swapping methods~\cite{fs}, whereas face-reenactment utilizes the reenactment technology to swap the expressions of a source video to a target video while maintaining the identity of the target person. 
In addition to the face-swapping forgeries above, entire image synthesis utilizes generative models such as GAN~\cite{karras2017progressive,karras2019style} and Diffusion models~\cite{ho2020denoising,song2020denoising,rombach2022high} to generate whole synthesis facial images directly without face-swapping operations such as blending.
Our work specifically focuses on detecting face-swapping but also shows the potential to detect entire image synthesis.

\vspace{-10pt}

\paragraph{Deepfake Detectors toward Generalization} 
The task of deepfake detection grapples profoundly with the issue of generalization. 
Recent endeavors can be classified into the detection of image forgery and video forgery.
The field of detecting image forgery have developed novel solutions from different directions: data augmentation~\cite{li2018exposing,li2020face,zhao2021learning,chen2022self,shiohara2022detecting}, frequency clues~\cite{qian2020thinking,liu2021spatial,luo2021generalizing,gu2022exploiting,wang2023dynamic}, ID information~\cite{dong2023implicit,huang2023implicit}, disentanglement learning~\cite{yang2021learning,liang2022exploring,yan2023ucf}, designed networks~\cite{dang2020detection,zhao2021multi}, reconstruction learning~\cite{wang2021representative,cao2022end}, and 3D decomposition~\cite{zhu2021face}. 
More recently, several works~\cite{jia2024can,tan2024data} attempt to generalize deepfakes with the designed training-free pipelines.
On the other hand, recent works of detecting video forgery focus on the temporal inconsistency~\cite{haliassos2021lips,zheng2021exploring,wang2023altfreezing}, eye blinking~\cite{li2018ictu}, 
landmark geometric features~\cite{sun2021improving}, neuron behaviors~\cite{wang2019fakespotter}, optical flow~\cite{amerini2019deepfake}.

\vspace{-10pt}

\paragraph{Deepfake Detectors Based on Data Augmentation}
One effective approach in deepfake detection is the utilization of data augmentation, which involves training models using synthetic data. 
For instance, in the early stages, FWA~\cite{li2018exposing} employs a self-blending strategy by applying image transformations (\textit{e.g.,} down-sampling) to the facial region and then warping it back into the original image. This process is designed to learn the wrapping artifacts during the deepfake generation process. 
Another noteworthy contribution is Face X-ray~\cite{li2020face}, which explicitly encourages detectors to learn the blending boundaries of fake images. 
Similarly, I2G~\cite{zhao2021learning} uses a similar method of Face X-ray to generate synthetic data and then employs a pair-wise self-consistency learning technique to detect inconsistencies within fake images.
Furthermore, SLADD~\cite{chen2022self} introduces an adversarial method to dynamically generate the most challenging blending choices for synthesizing data.
Rather than swapping faces between two different identities, a recent art, SBI~\citep{shiohara2022detecting}, proposes to swap with the same person's identity to reach a high-realistic face-swapping.


\begin{figure*}[!t] 
\centering 
\includegraphics[width=0.9\textwidth]{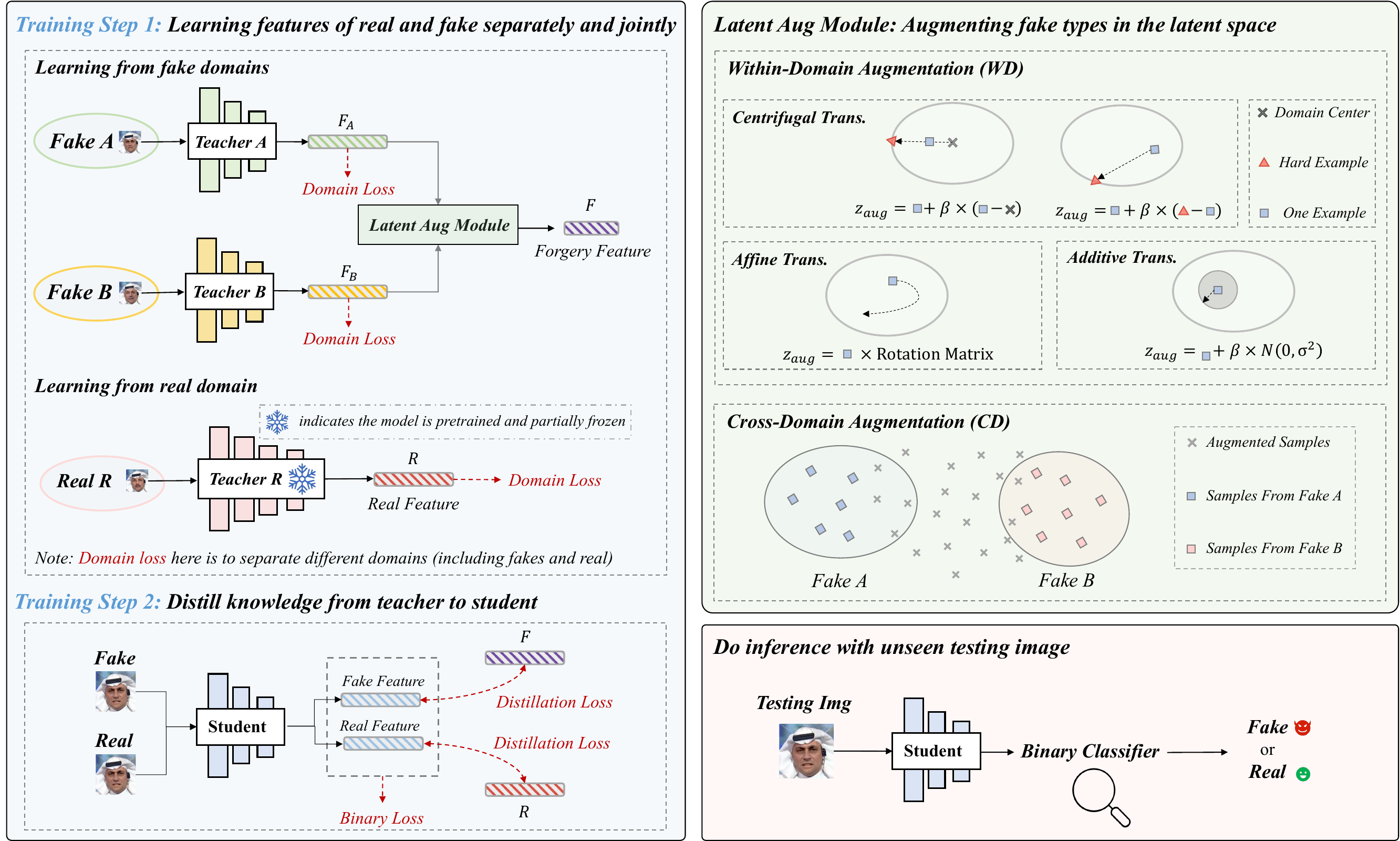} 
\caption{
The overall pipeline of our proposed method (two fake types are considered as an example). 
\textbf{(1)} In the training phase, the student encoder is trained to learn a generalizable and robust feature by utilizing the distribution match to distill the knowledge of the real and fake teacher encoders to the student encoder. 
\textbf{(2)} In the inference phase, only the student encoder is applied to detect the fakes from the real.
\textbf{(3)} For the learning of the forgery feature, we apply the latent space within-domain (WD) and cross-domain (CD) augmentation.
\textbf{(4)} For the learning of the real feature, the pre-trained and frozen ArcFace face recognition model is applied.
\textbf{(5)} WD involves novel augmentations to fine-tune domain-specific features, while CD enables the model to seamlessly identify transitions between different types of forgeries. 
} 
\label{fig:pipeline} 
\end{figure*}
\section{Method}


\subsection{Architecture Summary}
Our framework follows a novel \textbf{distillation-based learning architecture} beyond previous methods that train all data in a unique architecture. Our architecture consists of the teacher and student modules.
\textbf{Teacher module} involves: (1) Assigning a \underline{dedicated teacher encoder} to learn domain-specific features for each forgery type; (2) Applying within- and cross-domain augmentations to augment the forgery types; (3) Employing a \underline{fusion layer} to combine and fuse the features with the augmented.
\textbf{Student module} contains a single \underline{student encoder} with an FC layer. This encoder benefits from the learned features of the teacher module.


\subsection{Training Procedure}
\label{procedure}
The overall training process is summarized in Fig.~\ref{fig:pipeline}.
In the proposed framework, fake and real features are separately learned using distinct teacher encoders, facilitated by the \textbf{domain loss} (see ``Training Step 1" in Fig.~\ref{fig:pipeline}). 
In this step, the \textbf{latent augmentation module} is applied to augment the forgery types.
Subsequently, the learned features from both real and fake teacher encoders are combined to distill a student encoder with a binary classifier, guided by the \textbf{distillation loss} (see ``Training Step 2" in Fig.~\ref{fig:pipeline}). This student encoder is then encouraged to detect deepfakes (via the \textbf{binary loss}) using the features acquired from the teachers. During the whole training process, all teacher and student encoders are \textbf{trained jointly in an end-to-end manner.}
The rationale is that we aim to perform latent augmentation only within the forgery space.
By maintaining this separation, we aim to avoid the unintended combination of features from both real and fake instances. This approach aligns with our objective of expanding the forgery space without introducing real features.

\subsection{Latent Space Augmentation}
\label{lsa}
Suppose that we have a training dataset $\mathcal{D}=\bigcup_{i=0}^m d_i$, which contains $m$ type forgery images $\bigcup_{i=1}^m d_i$ and corresponding real type images $d_0$. First, we sample a batch of identities (face identities) and collect their image from each type of the dataset $\mathcal{D}$, where $\{\mathbf{x}_i \in \mathbb{R}^{B\times H\times W\times3}|\mathbf{x}_i\in d_i,i=0,1,...,m\}$. After inputting different types of images into the corresponding teacher encoder $f_i$, we perform our proposed latent space augmentation on the features $\mathbf{z}_i = f_i(\mathbf{x}_i)$, where $\mathbf{z}_i\in \mathbb{R}^{B \times C \times h \times w}$ and $i=0,1,...,m$.

As depicted in Fig.~\ref{fig:pipeline}, there are three different within-domain transformations, including the Centrifugal transformation (CT), Additive transformation (AdT), Affine transformation (AfT), and the cross-domain transformation.
We will introduce these augmentation methods as follows.

\subsubsection{Within-domain Augmentation}
The within-domain augmentation (WD) contains three specific techniques: centrifugal, affine, and additive transformations.
The Centrifugal transformation serves to create hard examples (far away from the centroid) that could encourage models to learn a more general decision boundary, as also indicated in \cite{shiohara2022detecting}. 
The latter two transformations are designed to help models learn a more robust representation by adding different perturbations. 

\paragraph{Centrifugal Transformation}
We argue that incorporating challenging examples effectively enlarges the space within each forgery domain.
Challenging examples, in this context, refer to samples that are situated far from the domain centroid. 
Therefore, transforming samples into challenging examples is to drive them away from the domain centroid $\boldsymbol{\mu}_i\in\mathbb{R}^{C\times h \times w}$, which can be computed by
\begin{equation}
   \boldsymbol{\mu}_i = \frac{1}{B}\sum_{j=1}^B (\mathbf{z}_i)_j, i=1,...,m,
\end{equation}
where $(\mathbf{z}_i)_j \in \mathbb{R}^{C \times h \times w}$ represents the $j$-th identity features within the batch $B$ of domain $i$.
We propose two kinds of augmentation methods that achieve our purpose in a \textbf{direct} and \textbf{indirect} manner, respectively. 
\begin{itemize}
\item \textbf{Direct manner}: We force $\mathbf{z}_i$ to move along the centrifugal direction as follows:
\begin{equation} \label{centrifugaldirect}
    \hat{\mathbf{z}_i} = \mathbf{z}_i + \beta (\mathbf{z}_i - \boldsymbol{\mu}_i), i=1,...,m,
\end{equation}
where $\beta$ is a scaling factor randomly sampled between 0 and 1. 
\item \textbf{Indirect manner}: We push $\mathbf{z}_i$ towards existing hard examples $\mathbf{a}_i\in\mathbb{R}^{C\times h \times w}$, the sample with the largest Euclidean distance from the center $\boldsymbol{\mu}_i$. We then transform $\mathbf{z}_i$ move towards hard examples by:
\begin{equation} \label{centrifugalindirect}
    \hat{\mathbf{z}_i} = \mathbf{z}_i + \beta (\mathbf{a}_i - \mathbf{z}_i), i=1,...,m.
\end{equation}
Here, $\beta$ is a scaling factor randomly sampled between 0 and 1.
\end{itemize}

\paragraph{Affine Transformation}
Affine transformation is proposed to transform the element-wise position information, creating neighboring samples.
Specifically, when we perform an affine rotation on $\mathbf{z}_i$ with rotation angle $\boldsymbol{\theta}$ in radians, we can derive the corresponding affine rotation matrix $\mathbf{A}$ as:
\begin{equation} \label{rotationmatrix}
    \mathbf{A} = \begin{bmatrix}
    \cos(\boldsymbol{\theta}) & -\sin(\boldsymbol{\theta}) & 0 \\
    \sin(\boldsymbol{\theta}) & \cos(\boldsymbol{\theta}) & 0 \\
    0 & 0 & 1
    \end{bmatrix}.
\end{equation}
After multiplying $\mathbf{A}$ with $\mathbf{P}$, the position information of $\mathbf{z}_i$ (\textit{i.e.,} the coordinate of each element in $\mathbf{z}_i$), the rotated position information $\hat{\mathbf{P}}$ is given by $\hat{\mathbf{P}} = \mathbf{A}\mathbf{P}$.
Then, we can obtain the rotated feature $\hat{\mathbf{z}_i}$ by rearranging elements' positions according to $\hat{\mathbf{P}}$.

\paragraph{Additive Transformation}
Adding perturbation is a traditional and effective augmentation, we apply this technique in latent space. 
By adding random noise, for example, Gaussian Mixture Model noise with zero mean, $\mathbf{z}_i$ can be perturbed with the scaling factor $\beta$ as follows:
\begin{equation} \label{additivetrans}
    \hat{\mathbf{z}_i} = \mathbf{z}_i + \beta \boldsymbol{\epsilon},
\end{equation}
where $\boldsymbol{\epsilon} \thicksim \sum_{k=1}^G \pi_k \mathcal{N}(\boldsymbol{\epsilon}|0, \boldsymbol{\Sigma}_k)$ and $\sum_{k=1}^G \pi_k=1$.

\subsubsection{Cross-domain Augmentation}
To create and interpolate the variants between different forgery domains, we utilize the Mixup augmentation technique~\cite{zhang2017mixup} in the latent space for cross-domain augmentation. This approach encourages the model to learn a more robust decision boundary and capture the general features shared across various forgeries. 
Specifically, we compute a linear combination of two latent representations: $\mathbf{z}_i$ and $\mathbf{z}_k$ that belong to different fake domains ($i\neq k$). The weight between two features is controlled by $\alpha$, which is randomly sampled between 0 and 1. The augmentation can be formally expressed as:
\begin{equation} \label{crosstransform}
    \hat{\mathbf{z}_i}^c = \alpha \mathbf{z}_i + (1 - \alpha) \mathbf{z}_k,i\neq k \in \{1,...,m\},
\end{equation}
where $i$ and $k$ are distinct forgery domains and $\hat{\mathbf{z}_i}^c$ stands for cross-domain augmented samples.

\subsubsection{Fusion layer}
Within each mini-batch, we perform both within-domain and cross-domain augmentation on $\mathbf{z}_i$ and obtain corresponding augmented representation $\hat{\mathbf{z}_i}\in \mathbb{R}^{B\times C \times h \times w}$ and $\hat{\mathbf{z}_i}^c\in \mathbb{R}^{B\times C \times h \times w}$, respectively. 
Then, we apply a learnable convolutional layer to bring augmentation results together to align the shape with the output of the student encoder:
\begin{equation} \label{finalaug}
    \hat{\mathbf{z}}_i^{aug} = \textit{Conv}(\hat{\mathbf{z}_i} \parallel  \hat{\mathbf{z}_i}^c), i=1,...,m,
\end{equation}
where $\parallel$ represents the concatenation operation along the channel dimension. 
Thus the final latent representation $\mathbf{F}_i\in \mathbb{R}^{B \times C\times h \times w}$ of forgery augmentation can be obtained by combining the original forgery representations and the augmented representations:
\begin{equation} \label{finalforgery}
    \mathbf{F}_i = \textit{Conv}(\hat{\mathbf{z}}_i^{aug} \parallel  \hat{\mathbf{z}_i}), i=1,...,m.
\end{equation}

\subsection{Objective Function}
\label{objective}

\paragraph{Domain Loss}
The domain loss is designed to encourage teacher encoders to learn domain-specific features (with each forgery type and the real category considered as distinct domains). After teacher encoders compress images $\mathbf{x}_i\in \mathbb{R}^{B \times H \times W \times 3}$ to $\mathbf{z}_i\in \mathbb{R}^{B\times C \times h \times w}$ in the latent space, we apply a multi-class classifier to estimate the confidence score $\mathbf{s}_i\in \mathbb{R}^{B\times (m+1)}$ that the feature is recognized as each domain.
The domain loss, given as a multi-class classification loss, can be represented by the Cross-Entropy Loss. At first, we turn the confidence score $\mathbf{s}_i$ into the likelihood $\mathbf{p}_i\in \mathbb{R}^B$: after the softmax, taking the $i$-th result, which is formulated as $\mathbf{p}_i = \text{softmax}(\mathbf{s}_i)[i]$. Then we compute the domain loss as follows:
\begin{equation} \label{domainloss}
\begin{aligned} 
    &\mathcal{L}_{domain} = - \frac{1}{B\times (m+1)} \times \\ 
    &\sum_{j=1}^B \left[ \log(1-(\mathbf{p}_0)_j) + \sum_{i=1}^{m} \log((\mathbf{p}_i)_j) \right],
\end{aligned}
\end{equation}
where $(\mathbf{p}_i)_j \in \mathbb{R}$ represents the forgery probability of $j$-th identity features within the batch $B$ of domain $i$ (0 is the real type).

\paragraph{Distillation Loss}
The distillation loss is the key loss to improve the generalization ability of the inference model by transferring augmented knowledge to the student: align the student's feature $\mathbf{F}_i^s$ with augmented latent representation $\mathbf{F}_i$. This alignment process is quantified using a distance measurement function $M(\cdot)$, which is formally as:

\begin{equation} \label{distill}
    \mathcal{L}_{distill} = \sum_{i=0}^m M(\mathbf{F}_i,\mathbf{F}_i^s).
\end{equation}

In the context of fake samples, the goal is to adjust the student model's feature map \( \mathbf{F}_i^s, i=1,...,m \) to approximate the comprehensive forgery representation \( \mathbf{F}_i, i=1,...,m \), where \( \mathbf{F}_i \) is obtained by Eq.~(\ref{finalforgery}).
Similarly, we align the student's feature map of the real \( \mathbf{F}_0^s \) to the teacher's real representation \( \mathbf{F}_0 \), where \( \mathbf{F}_0 \) is obtained by utilizing the pre-trained ArcFace~\cite{deng2019arcface} model.



\paragraph{Binary Classification Loss}
To finally achieve the Deepfake detection task, we add a binary classifier to the student encoder for detecting fakes from the real. The binary classification loss, commonly known as Binary Cross-Entropy, is formulated as follows:
\begin{equation}
\begin{aligned}
    &\mathcal{L}_{binary} = -\frac{1}{B\times (m+1)}\times
    \\ & \sum_{j=1}^B \left[ \log(1 - (\mathbf{p}_0)_j) + \sum_{i=1}^m \log((\mathbf{p}_i)_j) \right].
\end{aligned}
\end{equation}
In this equation, \( B \) represents the batch size of observations, and \( \mathbf{p}_i\) is the predicted probability that observation \( \mathbf{x}_i \) belongs to the class indicative of a deepfake, where $i=0,1,...,m$.

\paragraph{Overall Loss}
The final loss function is obtained by the weighted sum of the above loss functions.
\begin{equation} \label{overall_loss}
\mathcal{L} = \lambda_{1}\mathcal{L}_{binary} + \lambda_{2}\mathcal{L}_{domain} + \lambda_{3}\mathcal{L}_{distill},
\end{equation}
where $\lambda_{1}$, $\lambda_{2}$, and $\lambda_{3}$ are hyper-parameters for balancing the overall loss.

\begin{table*}
\centering
\scalebox{0.9}{
\begin{tabular}{c|c|c|c|c|c|c|c|c}
\toprule
\textbf{Method} & \textbf{Detector} & \textbf{Backbone} & \textbf{CDF-v1} & \textbf{CDF-v2} & \textbf{DFD} & \textbf{DFDC} & \textbf{DFDCP} & \textbf{Avg.} \\
\midrule
\midrule
Naive & Meso4~\cite{afchar2018mesonet} & MesoNet & 0.736 & 0.609 & 0.548 & 0.556 & 0.599 & 0.610 \\
Naive & MesoIncep~\cite{afchar2018mesonet} & MesoNet & 0.737 & 0.697 & 0.607 & 0.623 & 0.756 & 0.684 \\
Naive & CNN-Aug~\cite{he2016deep} & ResNet & 0.742 & 0.703 & 0.646 & 0.636 & 0.617 & 0.669 \\
Naive & Xception~\cite{rossler2019faceforensics++} & Xception & 0.779 & 0.737 & \underline{0.816} & 0.708 & 0.737 & 0.755 \\
Naive & EfficientB4~\cite{tan2019efficientnet} & EfficientNet & 0.791 & 0.749 & 0.815 & 0.696 & 0.728 & 0.756 \\
\midrule
Spatial & CapsuleNet~\cite{nguyen2019capsule} & Capsule & 0.791 & 0.747 & 0.684 & 0.647 & 0.657 & 0.705 \\
Spatial & FWA~\cite{li2018exposing} & Xception & 0.790 & 0.668 & 0.740 & 0.613 & 0.638 & 0.690 \\
Spatial & Face X-ray~\cite{li2020face} & HRNet & 0.709 & 0.679 & 0.766 & 0.633 & 0.694 & 0.696 \\
Spatial & FFD~\cite{dang2020detection} & Xception & 0.784 & 0.744 & 0.802 & 0.703 & 0.743 & 0.755 \\
Spatial & CORE~\cite{ni2022core} & Xception & 0.780 & 0.743 & 0.802 & 0.705 & 0.734 & 0.753 \\
Spatial & Recce~\cite{cao2022end} & Designed & 0.768 & 0.732 & 0.812 & 0.713 & 0.734 & 0.752 \\
Spatial & UCF~\cite{yan2023ucf} & Xception & 0.779 & 0.753 & 0.807 & \underline{0.719} & \underline{0.759} & 0.763 \\
\midrule
Frequency & F3Net~\cite{qian2020thinking} & Xception & 0.777 & 0.735 & 0.798 & 0.702 & 0.735 & 0.749 \\
Frequency & SPSL~\cite{liu2021spatial} & Xception & \underline{0.815} & \underline{0.765} & 0.812 & 0.704 & 0.741 & \underline{0.767} \\
Frequency & SRM~\cite{luo2021generalizing} & Xception & 0.793 & 0.755 & 0.812 & 0.700 & 0.741 & 0.760 \\
\midrule
\multirow{2}*{Ours} & \multirow{2}*{EFNB4 + LSDA} & \multirow{2}*{EfficientNet} & \textbf{0.867} & \textbf{0.830} & \textbf{0.880} & \textbf{0.736} & \textbf{0.815} & \textbf{0.826} \\
& & & \textcolor{cvprgreen}{(↑5.2\%)} & \textcolor{cvprgreen}{(↑6.5\%)} & \textcolor{cvprgreen}{(↑6.4\%)} & \textcolor{cvprgreen}{(↑1.7\%)} & \textcolor{cvprgreen}{(↑5.6\%)} & \textcolor{cvprgreen}{(↑5.9\%)} \\
\bottomrule
\end{tabular}
}
\caption{Cross-dataset evaluations using the \textbf{frame-level AUC} metric on the deepfake benchmark~\cite{yan2023deepfakebench}. All detectors are trained on FF++\_c23~\cite{rossler2019faceforensics++} and evaluated on other datasets. The best results are highlighted in bold and the second is underlined.}
\label{tab:table1}
\end{table*}

\section{Experiments}

\begin{table}[tb!]
  \centering
  \scalebox{0.9}{
  \begin{tabular}{c|c|c|c} \toprule
    Model& Publication & CDF-v2 & DFDC\\ 
    \midrule
    LipForensics~\cite{haliassos2021lips} & CVPR'21 & 0.824 & 0.735\\
    FTCN~\cite{zheng2021exploring} & ICCV'21 & 0.869 & 0.740 \\
    PCL+I2G~\cite{zhao2021learning} & ICCV'21 & 0.900 & 0.744 \\
    HCIL~\cite{gu2022hierarchical} & ECCV'22 & 0.790 & 0.692 \\
    RealForensics~\cite{haliassos2022leveraging} & CVPR'22 & 0.857 & \underline{0.759} \\
    ICT~\cite{dong2022protecting} & CVPR'22 & 0.857 & - \\
    SBI*~\cite{shiohara2022detecting} & CVPR'22 & \underline{0.906} & 0.724 \\
    AltFreezing~\cite{wang2023altfreezing} & CVPR'23 & 0.895 & - \\

    \midrule
    
    \multirow{2}*{Ours} & \multirow{2}*{-} & \textbf{0.911} & \textbf{0.770}\\

    & & \textcolor{cvprgreen}{(↑0.05\%)} & \textcolor{cvprgreen}{(↑1.1\%)} \\
    
    \bottomrule
  \end{tabular}
  }
  \caption{
  Comparison with recent state-of-the-art methods on CDF-v2 and DFDC using the \textbf{video-level AUC}.
  We report the results directly from the original papers.
  All methods are trained on FF++\_c23. * denotes our reproduction with the official code.
  The best results are in bold and the second is underlined.
  }
  \label{tab:cmp_sota}
  \vspace{-1pt}
\end{table}

\subsection{Settings}

\paragraph{Datasets.}
To evaluate the generalization ability of the proposed framework, our experiments are conducted on several commonly used deepfake datasets: FaceForensics++ (FF++)~\cite{rossler2019faceforensics++}, DeepfakeDetection (DFD)~\cite{dfd}, Deepfake Detection Challenge (DFDC)~\cite{dfdc}, preview version of DFDC (DFDCP)~\cite{dfdcp}, and CelebDF (CDF)~\cite{li2019celeb}. FF++~\cite{rossler2019faceforensics++} is a large-scale database comprising more than 1.8 million forged images from 1000 pristine videos. Forged images are generated by four face manipulation algorithms using the same set of pristine videos, \textit{i.e.,} DeepFakes (DF)~\cite{df}, Face2Face (F2F)~\cite{thies2016face2face}, FaceSwap (FS)~\cite{fs}, and NeuralTexture (NT)~\cite{thies2019deferred}. 
Note that there are three versions of FF++ in terms of compression level, \ie, raw, lightly compressed (c23), and heavily compressed (c40). Following previous works~\cite{li2020face,chen2022self,chen2022ost}, the c23 version of FF++ is adopted. 


\begin{table}[tb!]
  \centering
  \scalebox{0.8}{
  \begin{tabular}{c|c|c|c|c}
    \toprule
    \multirow{2}*{Method}
    &\multicolumn{4}{c}{Testing Datasets}\\
    \cmidrule(lr){2-5}
    & StarGAN~\cite{choi2018stargan} & DDPM~\cite{ho2020denoising} & DDIM~\cite{song2020denoising} & SD~\cite{rombach2022high} \\
    \midrule
    
    SBI~\cite{shiohara2022detecting} & 0.787 & 0.744 & 0.648 & 0.478 \\

    \midrule

    Ours & \textbf{0.810} &  \textbf{0.854}& \textbf{0.748} & \textbf{0.506} \\
     \bottomrule
  \end{tabular}}
  \caption{
  Results in detecting GAN-generated images and Diffusion-generated images. We compare our results with SBI~\cite{shiohara2022detecting}. We utilize its official code for evaluation. These models are trained on FF++\_c23. ``SD" is the short for stable diffusion.
  }
  \label{tab:detect_generative}
  \vspace{-1pt}
\end{table}

\begin{figure*}[!t] 
\centering 
\includegraphics[width=0.85\textwidth]{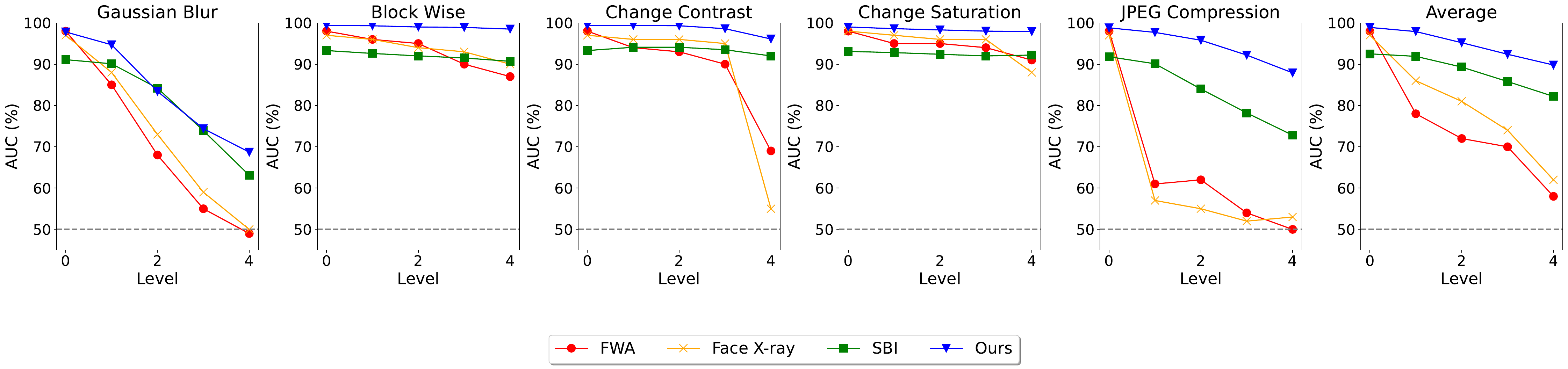} 
\caption{
Robustness to Unseen Perturbations: We report video-level AUC (\%) under five different degradation levels of five specific types of perturbations~\cite{jiang2020deeperforensics}. 
We compare our results with three RGB-based augmentation-based methods to demonstrate our robustness. Best viewed in color.
} 
\label{fig:dongge} 
\end{figure*}

\begin{table*}
  \centering
  \scalebox{0.69}{
  \begin{tabular}{c|c|c|c|c|c|c} \toprule
    \multirow{2}*{WD} & \multirow{2}*{CD} & CDF-v1 & CDF-v2 & DFDCP & DFDC & Avg.\\ 
    \cmidrule(lr){3-3}
    \cmidrule(lr){4-4}
    \cmidrule(lr){5-5}
    \cmidrule(lr){6-6}
    \cmidrule(lr){7-7}
    & & AUC  ~\textbar~  AP  ~\textbar~  EER 
    &  AUC  ~\textbar~  AP  ~\textbar~  EER 
    &  AUC  ~\textbar~  AP  ~\textbar~  EER 
    &  AUC  ~\textbar~  AP  ~\textbar~  EER 
    &  AUC  ~\textbar~  AP  ~\textbar~  EER \\
    \midrule
    $\times$ & $\times$ & 0.775 ~\textbar~ 0.843 ~\textbar~ 28.6 & 0.752 ~\textbar~ 0.847 ~\textbar~ 31.3 & 0.737 ~\textbar~ 0.846 ~\textbar~ 32.9 & 0.697 ~\textbar~ 0.721 ~\textbar~ 36.6 & 0.755 ~\textbar~ 0.846 ~\textbar~ 31.1 \\

   $\times$ & \checkmark & 0.862 ~\textbar~ 0.902 ~\textbar~ 21.1 & 0.819 ~\textbar~ 0.888 ~\textbar~ 26.0 & 0.807 ~\textbar~ 0.891 ~\textbar~ 27.6 & 0.733 ~\textbar~ 0.760 ~\textbar~ 33.5 & 0.821 ~\textbar~ 0.885 ~\textbar~ 25.5 \\

    \checkmark & $\times$ & \textbf{0.887} ~\textbar~ \textbf{0.925} ~\textbar~ \textbf{18.5} & \textbf{0.833} ~\textbar~ 0.903 ~\textbar~ \textbf{24.6} & 0.787 ~\textbar~ 0.869 ~\textbar~ 28.6 & 0.729 ~\textbar~ 0.750 ~\textbar~ 33.2 & 0.819 ~\textbar~ 0.885 ~\textbar~ 25.4 \\
    
    \checkmark & \checkmark & 0.867 ~\textbar~ 0.922 ~\textbar~ 21.9 & 0.830 ~\textbar~ \textbf{0.904} ~\textbar~ 25.9 & \textbf{0.815} ~\textbar~ \textbf{0.893} ~\textbar~ \textbf{26.9} & \textbf{0.736} ~\textbar~ \textbf{0.760} ~\textbar~ \textbf{33.0} & \textbf{0.825} ~\textbar~ \textbf{0.893} ~\textbar~ \textbf{25.5} \\    
    
    \bottomrule
  \end{tabular}
  }
  \caption{
  Ablation studies regarding the effectiveness of the within-domain (WD) and cross-domain (CD) augmentation strategies. All models are trained on the FF++\_c23 dataset and evaluated across various other datasets with metrics presented in the order of AUC~\textbar~AP~\textbar~EER (the frame-level). The average performance (Avg.) across all datasets are also reported. The best results are highlighted in bold.
  }
  \label{tab:ablation_wd_cd}
\end{table*}

\begin{table*}
  \centering
  \scalebox{0.69}{
  \begin{tabular}{c|c|c|c|c|c} \toprule
    \multirow{2}*{Real Encoder} & CDF-v1 & CDF-v2 & DFDCP & DFDC & Avg.\\ 
    \cmidrule(lr){2-2}
    \cmidrule(lr){3-3}
    \cmidrule(lr){4-4}
    \cmidrule(lr){5-5}
    \cmidrule(lr){6-6}
    &  AUC  ~\textbar~  AP  ~\textbar~  EER 
    &  AUC  ~\textbar~  AP  ~\textbar~  EER 
    &  AUC  ~\textbar~  AP  ~\textbar~  EER 
    &  AUC  ~\textbar~  AP  ~\textbar~  EER 
    &  AUC  ~\textbar~  AP  ~\textbar~  EER \\
    \midrule
    EFNB4~\cite{tan2019efficientnet} & 0.857 ~\textbar~ 0.908 ~\textbar~ 22.4 & 0.822 ~\textbar~ 0.893 ~\textbar~ 25.8 & 0.805 ~\textbar~ 0.885 ~\textbar~ 27.3 & 0.733 ~\textbar~ 0.759 ~\textbar~ 33.3 & 0.804 ~\textbar~ 0.861 ~\textbar~ 27.2 \\

    iResNet101~\cite{duta2021improved} & 0.854 ~\textbar~ 0.908 ~\textbar~ 23.0 & 0.792 ~\textbar~ 0.874 ~\textbar~ 28.1 & 0.797 ~\textbar~ 0.872 ~\textbar~ 27.2 & 0.715 ~\textbar~ 0.743 ~\textbar~ 35.7 & 0.790 ~\textbar~ 0.849 ~\textbar~ 28.5 \\

    ArcFace~\cite{deng2019arcface} & \textbf{0.867} ~\textbar~ \textbf{0.922} ~\textbar~ \textbf{21.9} & \textbf{0.830} ~\textbar~ \textbf{0.904} ~\textbar~ \textbf{25.9} & \textbf{0.815} ~\textbar~ \textbf{0.893} ~\textbar~ \textbf{26.9} & \textbf{0.736} ~\textbar~ \textbf{0.760} ~\textbar~ \textbf{33.0} & \textbf{0.812} ~\textbar~ \textbf{0.870} ~\textbar~ \textbf{26.9} \\ 
    
    \bottomrule
  \end{tabular}
  }
  \caption{
   Ablation studies regarding the effectiveness of the ArcFace pre-trained before the real encoder. The experimental settings are similar to Table.~\ref{tab:ablation_wd_cd}.
  }
  \label{tab:content_encoder}
  \vspace{-10pt}
\end{table*}

\vspace{-5pt}

\paragraph{Implementation Details.}
We employ EfficientNet-B4~\cite{tan2019efficientnet} as the default encoders to learn forgery features.
For the real encoder, we employ the model and pre-trained weights of ArcFace from the code\footnote{
\href{https://github.com/mapooon/BlendFace}{https://github.com/mapooon/BlendFace}.}. 
The model parameters are initialized through pre-training on the ImageNet. 
We also explore alternative network architectures and their respective results, which are presented in the \textbf{supplementary}.
We employ MSE loss as the feature alignment function ($M$ in eq.~(\ref{distill})).
Empirically, the $\lambda_1$, $\lambda_2$, and $\lambda_3$ are set to be 0.5, 1, and 1 in Eq.~(\ref{overall_loss}). We explore other variants in \textbf{supplementary}.
To ensure a fair comparison, all experiments are conducted within the DeepfakeBench~\cite{yan2023deepfakebench}. All of our experimental settings adhere to the default settings of the benchmark.
More details are in the \textbf{supplementary}.

\vspace{-10pt}

\paragraph{Evaluation Metrics.} 
By default, we report the \textbf{frame-level} Area Under Curve (AUC) metric to compare our proposed method with prior works.
Notably, to compare with other state-of-the-art detectors, especially the video-based methods, \textbf{we also report the video-level AUC to compare with.}
Other evaluation metrics such as Average Precision (AP) and Equal Error Rate (EER) are also reported for a more comprehensive evaluation.

\subsection{Generalization Evaluation}
All our experiments follow a commonly adopted generalization evaluation protocol by training the models on the FF++\_c23~\cite{rossler2019faceforensics++} and then evaluating on other previously untrained/unseen datasets (\eg, CDF~\cite{li2019celeb} and DFDC~\cite{dfdc}).

\vspace{-10pt}

\paragraph{Comparison with competing methods.}
We first conduct generalization evaluation on a unified benchmark (\ie, DeepfakeBench~\cite{yan2023deepfakebench}).
The rationale is that although many previous works have adopted the same datasets for training and testing, the pre-processing, experimental settings, \textit{etc}, employed in their experiments can vary. This variation makes it challenging to conduct fair comparisons.
Thus, we implement our method and report the results using DeepfakeBench~\cite{yan2023deepfakebench}.
For other competing detection methods, we directly cite the results in the DeepfakeBench and use the same settings in implementing our method for a fair comparison.
The results of the comparison between different methods are presented in Tab.~\ref{tab:table1}. It is evident that our method consistently outperforms other models across all tested scenarios. On average, our approach achieves a notable 5\% improvement in performance.

\vspace{-10pt}

\begin{figure}  
\centering 
    \includegraphics[width=0.38\textwidth]{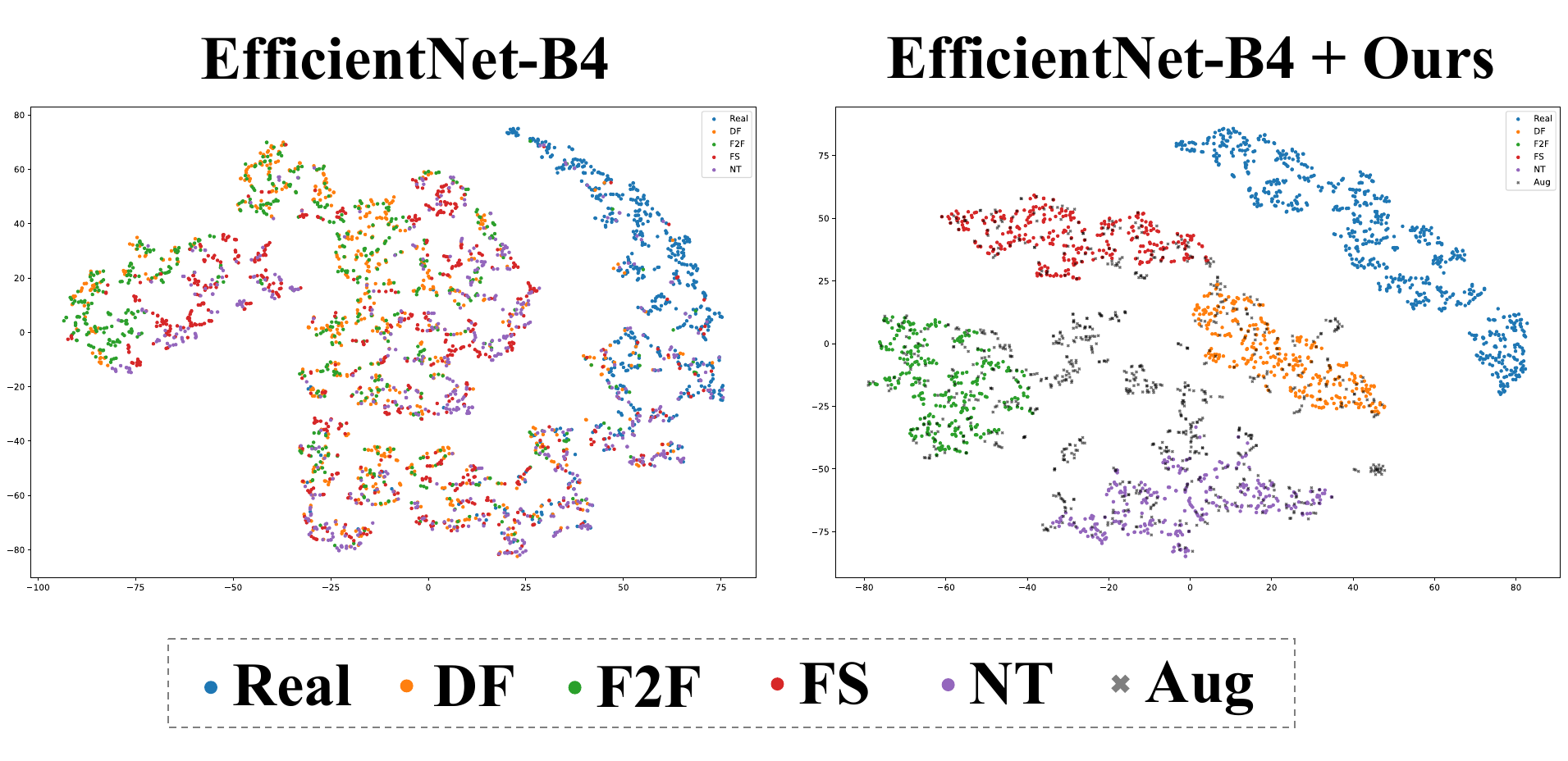} 
    \caption{t-SNE visualization of latent space \textit{w} and \textit{wo} augmentations.}
    \label{fig:tsne} 
\vspace{-4mm}
\end{figure}

\begin{figure}[!t] 
\centering 
\includegraphics[width=0.38\textwidth]{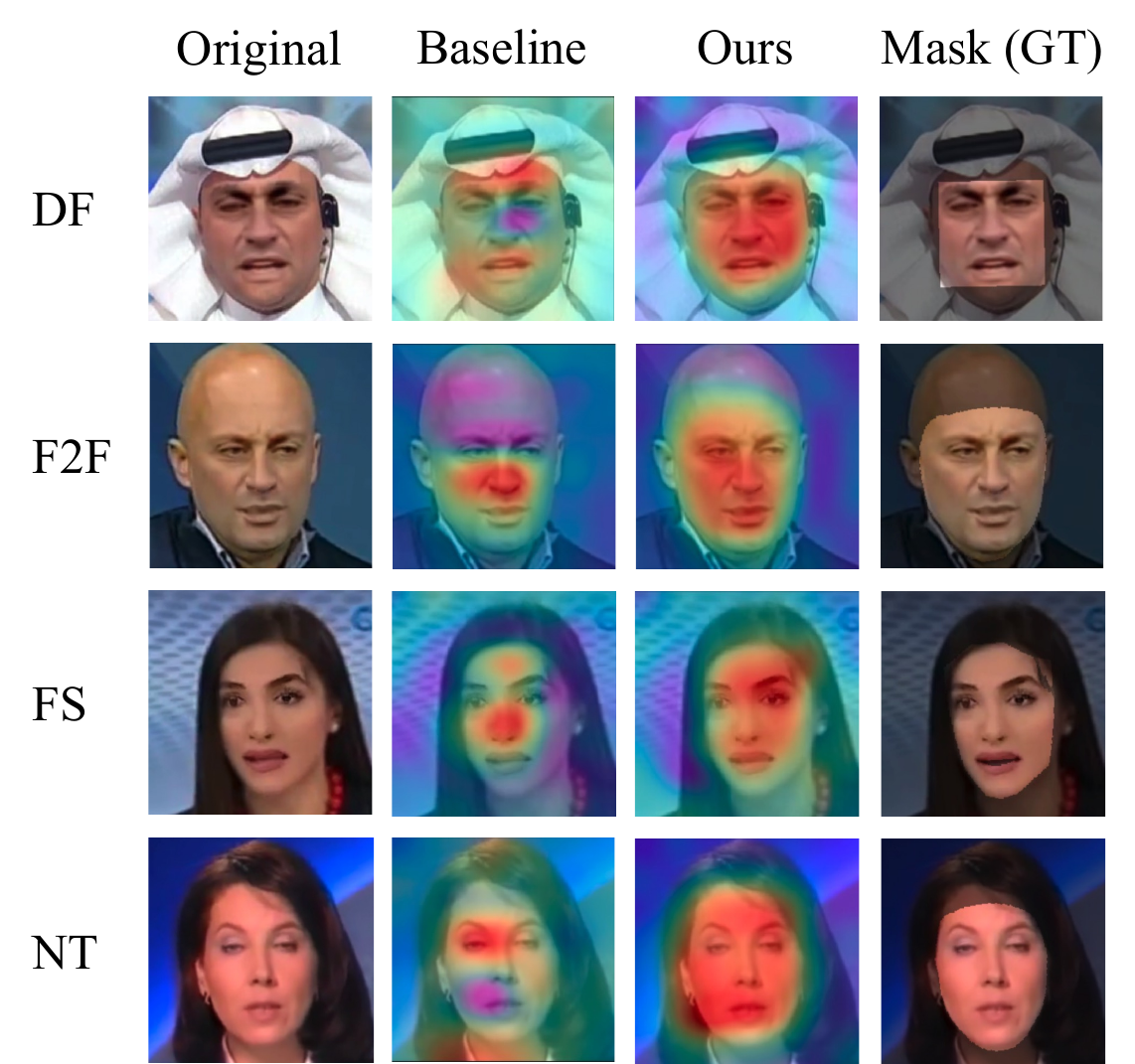} 
\caption{
GradCAM visualizations~\cite{selvaraju2017grad} for fake samples from different forgeries. We compare the baseline (EFNB4~\cite{tan2019efficientnet} with ours. ``Mask (GT)" highlights the ground truth of the manipulation region.
Best viewed in color.
} 
\label{fig:gradcam} 
\vspace{-15pt}
\end{figure}

\paragraph{Comparison with state-of-the-art methods.} 
In addition to the detectors implemented in DeepfakeBench, we further evaluate our method against other state-of-the-art models. We report the video-level AUC metric for comparison.
We select the recently advanced detectors for comparison, as listed in Tab.~\ref{tab:cmp_sota}.
%
%
Generally, the results are directly cited from their original papers. 
In the case of SBI, it is worth noting that the original results are obtained from training on the raw version of FF++, whereas other methods are trained on the c23 version. To ensure a fair and consistent comparison, we reproduce the results for SBI under the same conditions as the other methods.
The results, as shown in Tab.~\ref{tab:cmp_sota}, show the effective generalization of our method as it outperforms other methods, achieving the best performance on both CDF-v2 and DFDC.

\vspace{-10pt}

\paragraph{Comparison with RGB-based augmentation methods.}
To show the advantages of the latent space augmentation method (ours) over RGB-based augmentations (\eg, FWA~\cite{li2018exposing}, SBI~\cite{shiohara2022detecting}), we conduct several evaluations as follows.
\textbf{Robustness:}
RGB-based methods typically rely on subtle low-level artifacts at the pixel level. These artifacts could be sensitive to unseen random perturbations in real-world scenarios.
To assess the model's robustness to such perturbations, we follow the approach of previous works~\cite{haliassos2021lips}. 
Fig.~\ref{fig:dongge} presents the video-level AUC results for these unseen perturbations, utilizing the model trained on FF++\_c23. 
Notably, our method exhibits a significant performance advantage of robustness over other RGB-based methods.
\textbf{Extensibility}:
RGB-based methods classify an image as ``fake" if it contains evidence of a face-swapping operation, typically blending artifacts.
Beyond the evaluations on face-swapping datasets, we have extended our evaluation to include the detection in scenarios of entire face synthesis, which do not encompass blending artifacts.
For this evaluation, we compare our method SBI~\cite{shiohara2022detecting} that mainly relies on blending artifacts. 
The models are evaluated on both GAN-generated and Diffusion-generated data.
Remarkably, our method consistently outperforms SBI across all testing datasets (see Tab.~\ref{tab:detect_generative}). 
This observation shows the better extensibility of our detectors, which do not rely on specific artifacts like blending.



\begin{figure}[!t] 
\centering 
\includegraphics[width=0.38\textwidth]{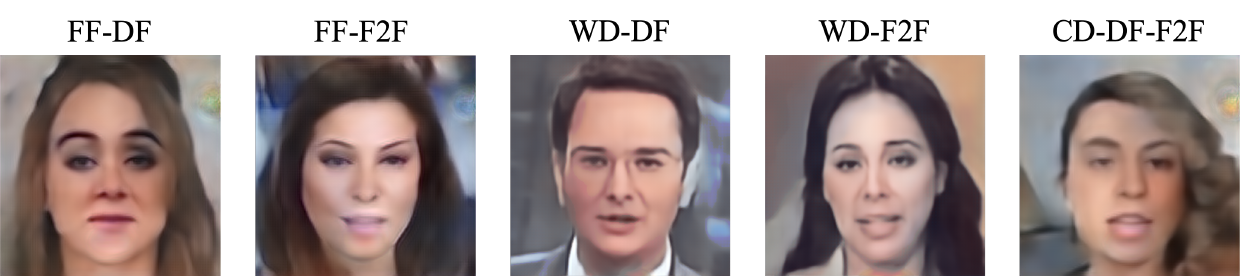} 
\caption{Visual examples of the original and augmented data.}
\label{fig:visual_example} 
\vspace{-10pt}
\end{figure}

\subsection{Ablation Study}

\paragraph{Effects of the latent space augmentation strategy.}
To evaluate the impact of the two proposed augmentation strategies (WD and CD), we conduct ablation studies on several datasets.
The evaluated variants include the baseline EfficientNet-B4, the baseline with the proposed within-domain augmentation (WD), the cross-domain augmentation (CD), and our overall framework (WD + CD).
The incremental enhancement in the overall generalization performance with the addition of each strategy, as evidenced by the results in Tab.~\ref{tab:ablation_wd_cd}, shows the effectiveness of these strategies. 
We also conduct ablation studies for each WD method in the \textbf{Supplementary}.

\vspace{-8pt}

\paragraph{Effects of face recognition prior.}
To assess the impact of the face recognition network (ArcFace~\cite{deng2019arcface}), we perform an ablation study comparing the results obtained using ArcFace (\textit{w} iResNet101 as the backbone) as the real encoder, to those achieved with the default backbone (\ie, EFNB4) and iResNet101 as the real encoder. As shown in Tab.~\ref{tab:content_encoder}, employing ArcFace as the real encoder results in notably better performance compared to using EFNB4 and iResNet101 (\textit{wo} face recognition pretraining) as the real encoder. This highlights the importance of utilizing the knowledge gained from face recognition, as offered by ArcFace, for deepfake detection tasks. Our findings align with those reported in our previous studies~\cite{haliassos2021lips,haliassos2022leveraging}.

\vspace{-5pt}

\section{Visualizations}

\paragraph{Visualizations of the captured artifacts.}
We further use GradCAM~\cite{zhou2016learning} to localize which regions are activated to detect forgery. 
The visualization results shown in Fig.~\ref{fig:gradcam} demonstrate that the baseline captures forgery-specific artifacts with a similar and limited area of response across different forgeries, while our model could locate the forgery region precisely and meaningfully.
In contrast, our method makes it discriminates between real and fake by focusing predominantly on the manipulated face area.
This visualization further identifies that LSDA encourages the baseline to capture more general forgery features.

\vspace{-8pt}

\paragraph{Visualizations of learned latent space.} 
We utilize t-SNE~\cite{van2008visualizing} for visualizing the feature space. We visualize the results on the FF++\_c23 testing datasets by randomly selecting 5000 samples.
Results in Fig.~\ref{fig:tsne} show our augmented method (the right) indeed learns a more robust decision boundary than the un-augmented baseline (the left).

\vspace{-3pt}

\section{Conclusion}
In this paper, we propose a simple yet effective detector that can generalize well in unseen deepfake datasets. 
Our key is that representations with a wider range of forgeries should learn a more adaptable decision boundary, thereby mitigating the overfitting to forgery-specific features. 
Following this idea, we propose to enlarge the forgery space by constructing and simulating variations within and across forgery features in the latent space.
Extensive experiments show that our method is superior in generalization and robustness to state-of-the-art methods.
We hope that our work will stimulate further research into the design of data augmentation in the deepfake detection community.

\vspace{-10pt}

\paragraph{Acknowledgment.}
Baoyuan Wu was supported by the National Natural Science Foundation of China under grant No.62076213, Shenzhen Science and Technology Program under grant No.RCYX20210609103057050, and the 
Longgang District Key Laboratory of Intelligent Digital Economy Security.
Qingshan Liu was supported by the National Natural Science Foundation of China under grant NSFC U21B2044.
Siwei Lyu was supported by U.S. National Science Foundation under grant SaTC-2153112.

\clearpage

{
    \small
    \bibliographystyle{ieeenat_fullname}
    \bibliography{main}
}


\end{document}